\documentclass[conference]{IEEEtran}
\IEEEoverridecommandlockouts

\usepackage[english]{babel}
\usepackage{cite}
\usepackage{amsmath,amssymb,amsfonts,amsthm}
\usepackage{algorithmic}
\usepackage{graphicx}
\usepackage{textcomp}
\usepackage{xcolor}
\usepackage{booktabs}
\usepackage{multirow}
\def\BibTeX{{\rm B\kern-.05em{\sc i\kern-.025em b}\kern-.08em
    T\kern-.1667em\lower.7ex\hbox{E}\kern-.125emX}}

\usepackage[utf8]{inputenc}
\newcommand{\mc}{\mathcal}
\newcommand{\R}{\mathbb{R}}
\newcommand{\expect}{\mathbb{E}}
\DeclareMathOperator{\var}{Var}
\newcommand{\dzvector}[1]{\mathbf{\boldsymbol{#1}}}
\renewcommand{\vec}[1]{\dzvector{#1}}
\DeclareMathOperator{\gdist}{d}
\newcommand{\frmu}{\expect^{\rm f}}
\newcommand{\frvar}{\var^{\rm f}}
\DeclareMathOperator*{\argmin}{arg\,min}
\newcommand{\norm}[1]{\left|{#1}\right|}
\theoremstyle{plain}

\begin{document}

\title{%
Autoregressive Models for Sequences of Graphs
\thanks{This research is funded by the Swiss National Science Foundation project 200021\_172671.
We gratefully acknowledge partial support of the Canada Research Chairs program.
We gratefully acknowledge the support of NVIDIA Corporation with the donation of the Titan Xp GPU used for this research.}
}

\author{%
\IEEEauthorblockN{Daniele Zambon\IEEEauthorrefmark{1}\thanks{\IEEEauthorrefmark{1}Equal contribution.}}
\IEEEauthorblockA{
    \textit{Università della Svizzera italiana},
    Lugano, Switzerland \\
    daniele.zambon@usi.ch
}
\\
\IEEEauthorblockN{Lorenzo Livi}
\IEEEauthorblockA{
    \textit{University of Manitoba}, 
    Winnipeg, Canada \\
    \textit{University of Exeter}, 
    Exeter, United Kingdom
}
\and
\IEEEauthorblockN{Daniele Grattarola\IEEEauthorrefmark{1}}
\IEEEauthorblockA{
    \textit{Università della Svizzera italiana},
    Lugano, Switzerland \\
    daniele.grattarola@usi.ch
}
\\
\IEEEauthorblockN{Cesare Alippi}
\IEEEauthorblockA{
    \textit{Università della Svizzera italiana}, Lugano, Switzerland \\
    \textit{Politecnico di Milano}, 
    Milano, Italy
}
}
\maketitle

\begin{abstract}
This paper proposes an autoregressive (AR) model for sequences of graphs, which generalises traditional AR models.
A first novelty consists in formalising the AR model for a very general family of graphs, characterised by a variable topology, and attributes associated with nodes and edges.
A graph neural network (GNN) is also proposed to learn the AR function associated with the graph-generating process (GGP), and subsequently predict the next graph in a sequence.
The proposed method is compared with four baselines on synthetic GGPs, denoting a significantly better performance on all considered problems.
\end{abstract}

\section{Introduction}
Several physical systems can be described by means of autoregressive (AR) models and their variants.
In that case, a system is represented as a discrete-time signal in which, at every time step, the observation is modelled as a realisation of a random variable that depends on preceding observations. 
In this paper, we consider the problem of designing AR predictive models where the observed entity is a graph.

In the traditional setting for AR predictive models, each observation generated by the process is modelled as a vector, intended as a realisation of random variable $x_{t+1}\in\R^d$, so that
\begin{equation}
    \label{eq:ar-multivariate}
    \begin{cases}
    x_{t+1} = f(x_t, x_{t-1},\dots,x_{t-p+1}) + \epsilon, \\ 
    \expect[\epsilon] = 0, \\
    \var[\epsilon]=\sigma^2 < \infty.
    \end{cases}
\end{equation}
In other terms, each observation $x_{t+1}$ is obtained from the regressor 
\[
    \vec x_t^p = [x_t, x_{t-1},\dots,x_{t-p+1}]
\]
through an AR function $f(\cdot)$ of order $p$ affected by an additive stationary random noise $\epsilon$. 
Given \eqref{eq:ar-multivariate}, the prediction of $x_{t+1}$ is often taken as  
\begin{equation}
    \label{eq:predictor-num}
    \hat x_{t+1} = \expect_\epsilon[x_{t+1}] =  f(x_t, x_{t-1},\dots,x_{t-p+1}),
\end{equation}
where the expectation $\expect_\epsilon[\cdot]$ is with respect to noise at time $t+1$. The predictor from \eqref{eq:predictor-num} is optimal when considering the $L^2$-norm loss between $\hat x_{t+1}$ and $x_{t+1}$.

Predicting the temporal evolution of a multivariate stochastic process is a widely explored problem in system identification and machine learning. 
However, several recent works have focused on problems where the vector representation of a system can be augmented by also considering the relations existing among variables \cite{battaglia2018relational}, ending up with a structured representation that can be naturally described by a graph. 
Such structured information provides an important inductive bias that the learning algorithms can take advantage of.
For instance, it is difficult to predict the chemical properties of a molecule by only looking at its atoms; on the other hand, by explicitly representing the chemical bonds, the description of the molecule becomes more complete, and the learning algorithm can take advantage of that information.
Similarly, several other machine learning problems benefit from a graph-based representation, e.g., the understanding of visual scenes \cite{raposo2017discovering}, the modelling of interactions in physical and multi-agent systems \cite{battaglia2016interaction,kipf2018neural}, or the prediction of traffic flow \cite{cui2018high}. 
In all these problems (and many others \cite{battaglia2018relational}), the dependencies among variables provide a strong prior that has been successfully leveraged to significantly surpass the previous state of the art.

In this paper, we consider attributed graphs where each node and edge can be associated with a feature vector. 
As such, a graph with $N$ nodes can be seen as a tuple $(\mc V, \mc E)$, where 
\begin{equation*}
    \label{eq:graph-nodes}
    \mc V = \{v_i \in \R^F\}_{i=1,\dots,N}
\end{equation*}
represents the set of nodes with $F$-dimensional attributes, and 
\begin{equation*}
    \mc E = \{e_{ij} \in \R^S\}_{v_i,v_j \in \mc V}
\end{equation*}
models the set of $S$-dimensional edge attributes \cite{gm_survey}. By this formulation also categorical attributes can be represented, via vector encoding.
We denote with $\mc G$ the set of all possible graphs with vector attributes.
Graphs in $\mc G$ can have different order and topology, as well as variable node and edge attributes. 
Moreover, a correspondence among the nodes of different graphs might not be present, or can be unknown; in this case, we say that the nodes are \emph{non-identified}.

Here we build on the classic AR model \eqref{eq:ar-multivariate} operating on discrete-time signals, and propose a generalised formulation that allows us to consider sequences of generic attributed graphs. 
The formulation is thus adapted by modelling a graph-generating process (GGP) that produces a sequence $\{g_1,g_2,\dots,g_t,\dots\}$ of graphs in $\mc G$, where each graph $g_{t+1}$ is obtained through an AR function 
\begin{equation}
    \label{eq:g-reg-fun}
    \phi(\cdot): \mc G^p \rightarrow \mc G.
\end{equation}
on the space $\mc G$ of graphs.
In order to formulate an AR model equivalent to \eqref{eq:ar-multivariate} for graphs, the AR function $f(\cdot)$ must be defined on the space of graphs, and the concept of additive noise must also be suitably adapted to the graph representation. 

Previous art on predictive autoregression in the space of graphs is scarce, with \cite{richard2014link} being the most notable contribution to the best of out knowledge. However, the work in \cite{richard2014link} only considers binary graphs governed by a vector AR process of order $p=1$ on some graph features, and does not allow to consider attributes or non-identified nodes.

The novel contribution of this work is therefore two-fold. 
First, we formulate an AR system model generating graph-valued signals as a generalisation of the numerical case. In this formulation, we consider graphs in a very general form. 
In particular, our model deals with graphs characterised by:
\begin{itemize}
    \item directed and undirected edges;
    \item identified and non-identified nodes;
    \item variable topology (i.e., connectivity);
    \item variable order;
    \item node and edge attributes (not necessarily numerical vectors, but also categorical data is possible).
\end{itemize}
Our model also accounts for the presence of an arbitrary stationary noise, which is formulated as a graph-valued random variable that can act on the topology, node attributes, and edge attributes of the graphs.

Second, we propose to learn the AR function \eqref{eq:g-reg-fun} using the recently proposed framework of graph neural networks (GNNs) \cite{battaglia2018relational}, a family of learnable functions that are designed to operate directly on arbitrary graphs.
This paper represents a first step towards modelling graph-valued stochastic processes, and introduces a possible way of applying existing tools for deep learning to this new class of prediction problems.

The rest of the paper is structured as follows: Section \ref{sec:graph-ar-model} formulates the problem of defining AR models in the space of graphs and Section \ref{sec:neural-graph-ar-function} introduces the proposed architecture for the GNN; Section \ref{sec:experiments} reports the experimental analysis performed to validate the methodology.

\section{Neural Graph Autoregression}
\label{sec:graph-ar-framework}

\subsection{Autoregressive Model for a GGP}
\label{sec:graph-ar-model}

Due to the lack of basic mathematical operators in the vast family of graphs we consider here, the generalisation of model \eqref{eq:ar-multivariate} to account for graph data is non-trivial. For instance, we have to deal with the fact that the sum between two graphs is not defined, although it can be formulated in some particular cases, e.g., when two graphs have the same order, identified nodes and numerical attributes \cite{beineke2004topics}.

Let $\{g_1, g_2,\dots, g_t,\dots\}$ be a discrete signal where each sample is described by a graph data structure.
As done for the numerical case, we model each observation of the process as a realisation of a random variable $g_{t+1} \in \mc G$, dependent on a graph-valued regressor
\[
    \vec g_t^{p} = [g_t, g_{t-1}, \dots, g_{t-p+1}],
\] 
through an AR function $\phi(\cdot)$ on the space of graphs.
Similarly to \eqref{eq:ar-multivariate}, $g_{t+1}$ is modelled as
\begin{equation}
\label{eq:graph_ar_model}
    g_{t+1} = H(\phi(\vec g_t^p),\eta),
\end{equation}
where 
\[
    H:\mc G \times \mc G \rightarrow \mc G.
\]
is a function that models the effects of noise graph%
\footnote{In the present paper, with noise graph we mean a graph-valued random variable distributed according to a stationary graph distribution \cite{zambon2017concept}.}
$\eta\in\mc G$ on graph $\phi(\vec g_t^p)$.
Function $H(\cdot,\cdot)$ in \eqref{eq:graph_ar_model} is necessary because, as mentioned above, the sum between graphs  $\phi(\vec g_t^p)$ and $\eta$ is not generally defined. 

Assumptions made  on the noise in \eqref{eq:ar-multivariate} have to be adapted as well. 
In the classic formulation \eqref{eq:ar-multivariate}, the condition of unbiased noise is formulated as $\expect[\epsilon]=0$ or, equivalently, as $f(x_{t},\dots, x_{t-p-1})=\expect_\epsilon[f(x_{t},\dots, x_{t-p-1})+\epsilon]$. 
In the space of graphs the assumption of unbiased noise can be extended as 
\begin{equation}
    \label{eq:no_bias_graph}
    \phi(\vec g_t^p) \in \frmu_{\eta}[H(\phi(\vec g_t^p),\eta)],
\end{equation}
where $\frmu[\cdot] \in \mc G$ is the set of mean graphs according to Fr\'echet \cite{frechet1948elements}, defined as the set of graphs minimising: 
\begin{equation}
\label{eq:frechet-mean}
    \frmu[g]= \argmin_{g'\in\mc G} \int_{\mc G} \gdist(g, g')^2 \,dQ_g(g).
\end{equation}
Function $\gdist(\cdot, \cdot)$ in \eqref{eq:frechet-mean} is a pre-metric distance between two graphs, and $Q_g$ is a graph distribution defined on the Borel sets of space $(\mc G,\gdist)$. 
Examples of graph distances $\gdist(\cdot,\cdot)$ that can be adopted are the graph edit distances \cite{abu2015exact,fankhauser2011speeding,bougleux2017graph}, or any distance derived by positive semi-definite kernel functions \cite{sejdinovic2013equivalence}. 
Depending on the graph distribution $Q_g$, there can be more than one graph minimising \eqref{eq:frechet-mean}, hence ending up with $\frmu[\cdot]$ as a set.
We stress that for a sufficiently small Fr\'echet variation of the noise graph $\eta$ (see Eq. \eqref{eq:frechet-var}) and a metric $\gdist(\cdot,\cdot)$, the Fr\'echet mean graph exists and is unique.

Equation \eqref{eq:frechet-mean} holds only when \[
    \int_{\mc G} \gdist(g, g')^2 \,dQ_g(g) < \infty,
\] 
which can be interpreted as the graph counterpart of $\var[\epsilon]<\infty$ in \eqref{eq:ar-multivariate}. The variance  in the graph domain can be expressed in terms of the Fr\'echet variation:
\begin{equation}
\label{eq:frechet-var}
    \frvar[g]:=\min_{g'\in\mc G} \int_{\mc G} \gdist(g, g')^2 \,dQ_g(g) < \infty.
\end{equation}

The final AR system model in the graph domain becomes
\begin{equation}
\label{eq:gar}
\begin{cases}
    g_{t+1} = H(\phi(\vec g_t^p),\eta), 
    \\\phi(\vec g_t^p)\in \frmu_{\eta}[H(\phi(\vec g_t^p),\eta)], 
    \\\frvar[\eta] < \infty.
\end{cases}
\end{equation}

Notice that the proposed graph autoregressive model~\eqref{eq:gar} is a proper generalisation of model \eqref{eq:ar-multivariate}. 
In fact, it can be shown that \eqref{eq:gar} reduces to  \eqref{eq:ar-multivariate}, when considering $(\R, \norm{\cdot})$ -- or more generally, $(\R^d, \norm{\cdot}_2)$ -- in place of $(\mc G,\gdist)$, and choosing $H(a,b)$ as the sum $a+b$ (see Appendix~\ref{app:remark:equivalence-proof} for a proof).

Given a GGP modelled by \eqref{eq:gar}, we can predict graph at time ${t+1}$ as that graph $\hat g_{t+1}$ minimising quantity
\[
\expect\left[\gdist\left(g, g_{t+1} \right)^2\right]
\]
where the expectation is taken with respect to ${g_{t+1}}$.
Therefore, we obtain that the optimal prediction is  attained at graph
\begin{equation}
\label{eq:optimal-g_t+1}
    \begin{array}{rl}
    \hat g_{t+1} &= \argmin_{g\in\mc G} \expect\left[\gdist\left(g, g_{t+1} \right)^2\right]
    \\ & 
    = \frmu_{\eta}\left[H\left(\phi(\vec g_t^p),\eta\right)\right]
    = \phi(\vec g_t^p).
    \end{array}
\end{equation}

\subsection{Learning the AR function with a Graph Neural Network}
\label{sec:neural-graph-ar-function}

\begin{figure*}
    \centering
    \includegraphics[width=.8\textwidth]{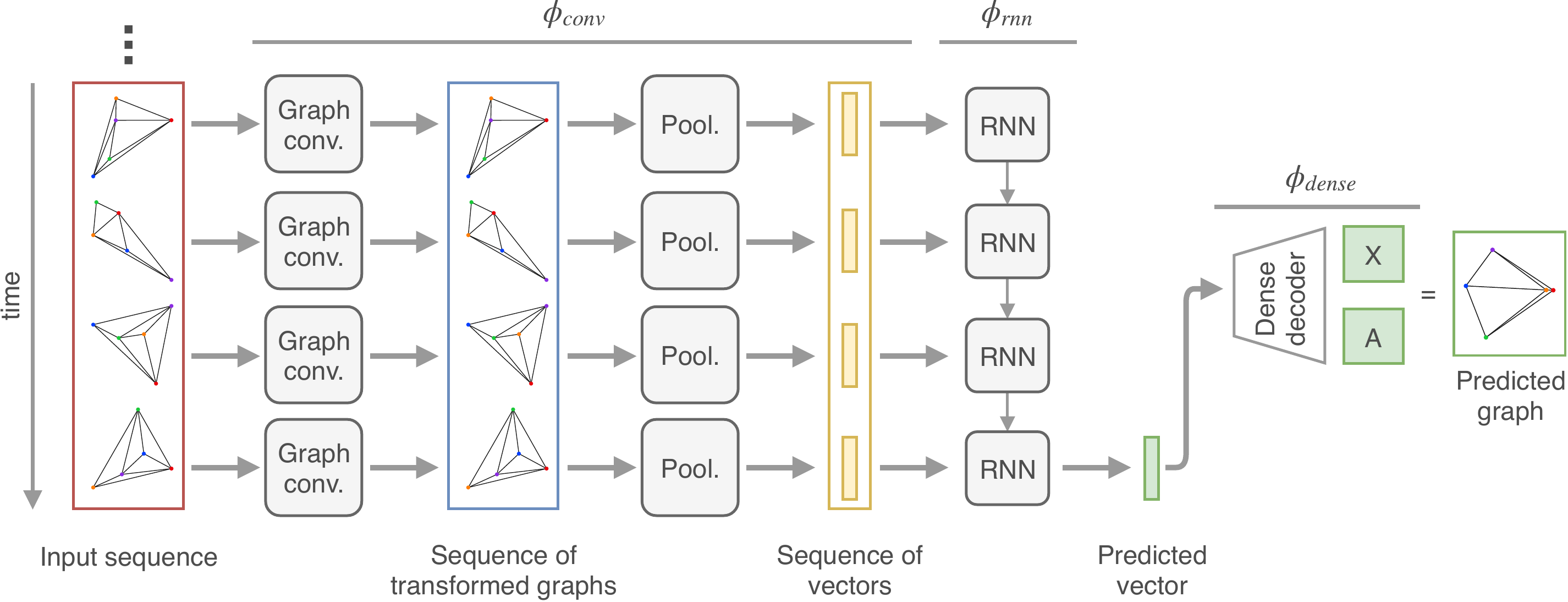}
    \caption{Schematic view of the proposed GNN. The network takes a sequence of $k$ graphs up to time $t$ as input (red box), and is trained to predict the graph at time $t+1$. The input sequence is processed by a graph convolutional network, which is applied in parallel to all graphs, followed by a global pooling layer \cite{li2015gated} that compresses the graphs down to vectors. The resulting sequence of vectors (yellow box) is processed by a recurrent neural network, which outputs a vector representing the predicted graph. Finally, the predicted vector is transformed in a graph by a multi-layer perceptron with two parallel outputs: one for producing the node features $X_{t+1}$, and one for the adjacency matrix $A_{t+1}$.}
    \label{fig:arch}
\end{figure*}

Given a GGP described by an AR model of order $p$, the task of predicting the next graph in the sequence can be formulated as that of approximating $\phi(\cdot)$ in \eqref{eq:gar}, as the optimal prediction is given by \eqref{eq:optimal-g_t+1}.
In order to approximate $\phi(\cdot)$ we propose to use a GNN,  which can be seen as a family of models
\begin{equation*}
    \phi_{nn}(\,\cdot\,;\theta): \mc G^k \rightarrow \mc G    
\end{equation*}
parametrised in vector $\theta$.
The family of models receives
regressor $\vec g_t^k=[g_{t},\dots,g_{t-k+1}]$ and outputs the predicted graph $\hat g_{t+1}$ 
\begin{equation}
    \hat g_{t+1}=\phi_{nn}(\vec g_t^k;\theta). 
\end{equation}
As the order $p$ is usually unknown, the value $k$ is a hyperparameter that must be appropriately chosen. 

We propose a possible architecture for $\phi_{nn}$, composed of three main blocks:
\begin{enumerate}
    \item each graph is mapped to an abstract vector representation, using graph convolutions \cite{gori2005new,scarselli2009graph,bruna2013spectral} and graph pooling \cite{defferrard2016convolutional,li2015gated} layers;
    \item a predictive model is applied to the resulting vector sequence;
    \item the predicted graph is obtained by mapping the predicted vector back to the graph domain.
\end{enumerate}
The full GNN $\phi_{nn}$ is therefore obtained by the composition of three blocks, denoted $\phi_{conv}$, $\phi_{rnn}$, and $\phi_{dense}$. 
Although each block has its own parameters $\theta_{conv}, \theta_{rnn} $  and $\theta_{dense}$, the network can be trained end-to-end as a single family of models with $\theta=\{\theta_{conv},\theta_{rnn},\theta_{dense}\}$. 
The three blocks are defined as follows (see also Figure \ref{fig:arch} for a schematic view of the architecture).

The first block of the network converts the input sequence $\vec g_t^k$ to a sequence of $l$-dimensional vectors. This operation can be described by map:
\begin{equation*}
    \phi_{conv}(\,\cdot\,; \theta_{conv}): \mc G \rightarrow \R^{l},
\end{equation*}
which is implemented as a GNN alternating graph convolutional layers for extracting local patterns, and graph pooling layers to eventually compress the representation down to a single vector.
By mapping graphs to a vector space, we go back to the numerical setting of \eqref{eq:ar-multivariate}. Moreover, the vector representation of the graphs takes into account the relational information that characterises the GGP.

By applying $\phi_{conv}$ to each graph in the regressor $\vec g_t^k$ (with the same $\theta_{conv}$), we obtain a sequence
\[
    [\phi_{conv}(g_t; \theta_{conv}),\dots, \phi_{conv}(g_{t-k+1}; \theta_{conv})]
\]
of $l$-dimensional vectors, which is then processed by block
\begin{equation*}
    \phi_{rnn}(\,\cdot\,;\theta_{rnn}): \R^{k \times l} \rightarrow \R^l.
\end{equation*}
The role of this block is to produce the vector representation of the predicted graph, while also capturing the temporal dependencies in the input sequences. Here, we formulate the block as a recurrent network, but any method to map the sequence to a prediction is suitable (e.g., fully connected networks). 

Finally, we convert the vector representation to the actual prediction in the space of graphs, using a multi-head dense network similar to the ones proposed in \cite{simonovsky2018graphvae,de2018molgan}:
\begin{equation*}
    \phi_{dense}(\,\cdot\,;\theta_{dense}): \R^l \rightarrow \mc G.
\end{equation*}

Note that generating a graph by sampling its topology and attributes with a dense network has known limitations and implicitly assumes node identity and a maximum order. While this solution is more than sufficient for the experiments conducted in this paper, and greatly simplifies the implementation of the GNN, other approaches can be used when dealing with more complex graphs, like the GraphRNN decoder proposed in \cite{you2018graphrnn}. 

\section{Experiments}
\label{sec:experiments}

The experimental section aims at showing that AR models for graphs are effective. 
In particular, we show that the proposed neural graph autoregressive (NGAR) model can be effectively trained, and that it provides graph predictions that significantly improve over simpler baselines.

The experiments are performed on sequences of attributed graphs with $N=5$ identified nodes and a variable topology. Each node is associated with a vector attribute of dimension $F=2$, and no edge attributes; however, the extension to graphs with edge attributes, as well as graphs of variable order and non-identified nodes, is straightforward (e.g., see \cite{simonovsky2018graphvae} for an example).
The sequences are produced by two synthetic GGPs that we generate with a controlled memory order $p$ (details follow in Section~\ref{sec:synth-ggp}), allowing us to have a ground truth for the analysis.

The comparative analysis of the tested methods is performed by considering a graph edit distance (GED) \cite{abu2015exact} 
\begin{equation}
\label{eq:ged-error}
    \gdist(g_{t+1},\hat g_{t+1})
\end{equation}
between the ground truth and the prediction made by the models%
\footnote{Although the NGAR method was trained with a specific loss function (see Sec.\ref{sec:ngar-arch}), here we considered the GED measure instead, in order to provide a fair comparison of the methods.}.
We also analyse the performance of NGAR in terms of prediction loss and accuracy, in order to show the relative performance of the model on problems of different complexity. Finally, we report a qualitative assessment of the predictions of NGAR, by visualising the graphs predicted by the GNN and the true observations from the GGP.

The rest of this section introduces the baselines (Section~\ref{sec:baselines}), the synthetic GGPs (Section~\ref{sec:synth-ggp}) and the implementation details for the NGAR architecture (Section~\ref{sec:ngar-arch}). Finally, we discuss the results of the experimental analysis in Section~\ref{sec:results}.

\subsection{Baseline methods}
\label{sec:baselines}
We consider four baselines commonly applied in the numerical case, that can be easily adapted to our setting. 
We denote our proposed method as \textit{NGAR}, while the four baselines as \textit{Mean}, \textit{Mart}, \textit{Move}, and \textit{VAR}, respectively.

\paragraph{Mean} 
the first baseline assumes that the GGP is stationary, with independent and identically distributed graphs. In this case, the optimal prediction $\hat g_{t+1}$ is the mean graph:
\begin{equation*}
    \hat g_{t+1} = \frmu[g], 
\end{equation*}
where $g\sim Q$, and $Q$ indicates here the distribution (supposed stationary) of the graphs.

\paragraph{Mart} 
the second baseline assumes that the GGP is a martingale, s.t.\ $\frmu[g_{t+1}]=g_t$, and predicts $g_{t+1}$ as: 
\begin{equation*}
    \hat g_{t+1} = g_{t},
\end{equation*} 
i.e., the graph at the previous time step.

\paragraph{Move} 
the third baseline considers the $k$ preceding graphs $\vec g_t^k$, and predicts $g_{t+1}$ to be the Fr\'echet sample mean graph of $\vec g_t^k$
%
\begin{equation*}
    \hat g_{t+1} = 
    \argmin\limits_{g'} \sum\limits_{g_i \in \vec g_t^k} \gdist(g', g_i)^2.
\end{equation*}

\paragraph{VAR} 
the fourth baseline is a vector AR model (VAR) of order $k$, which treats each graph $g_t$ as the vectorisation of its node features $X_t$ and adjacency matrix $A_t$, concatenated in a single vector:
\[
    u_t=[\text{vec}(X_t)^\top, \text{vec}(A_t)^\top]^\top \in \R^{N \cdot F+N^2}. 
\]
First, we compute a prediction $\hat u_{t+1}$ defined by the linear model
\[
    \hat u_{t+1} = B_0 + \sum_{i=1}^{k} B_i\cdot u_{t-i+1},
\]
from the regressor $\vec u_t^k =[u_t,\dots,u_{t-k+1}^k]$ and, subsequently, the actual graph prediction $\hat g_{t+1}$ is re-assembled from $\hat u_{t+1}$. 

We mention that baseline \emph{VAR} can be adopted in these experiments only because we are considering graphs with numerical attributes and fixed order; in more general settings this would not be possible. 

\subsection{Graph-sequence generation}
\label{sec:synth-ggp}
We consider two different GGPs, both based on a common framework where a multivariate AR model of the form
\begin{equation}
\label{eq:hidden-AR-model} 
    x_{t+1}= f(\vec x_t^p) + \epsilon,
\end{equation}
produces a sequence of vectors $\{x_1, x_2, \dots, x_t,\dots\}$, which are then used to generate the graph sequence. 
From each $x_t$, a node-feature matrix $X_t\in\R^{N \times F}$ is built, and the adjacency matrix $A_t$ is created as the Delaunay triangulation of the rows of $X_t$, interpreted as points in $\R^F$.
Here, $\epsilon$ is a noise term whose components are randomly drawn from a stationary normal distribution $\mc N(0,\sigma^2)$.

The result of this process is a sequence of graphs $\{g_1, g_2, \dots, g_t,\dots\}$, where each graph depends on the previous $p$ graphs. 
By choosing different implementations of $f(\vec x_t^p)$, we are able to generate different GGPs.
Note that the noise $\eta$ and function $H(\cdot,\cdot)$ in \eqref{eq:gar} are never made explicit in this procedure, but instead they are determined by the Gaussian noise perturbation introduced by $\epsilon$, which affects the node attributes and, consequently, causes possible changes in the topology.
Section~\ref{sec:rotational} and Section~\ref{sec:partial} describe the two procedures.

\subsubsection{Rotational model}
\label{sec:rotational}
\begin{figure}[t]
    \centering
    \includegraphics[width=\columnwidth]{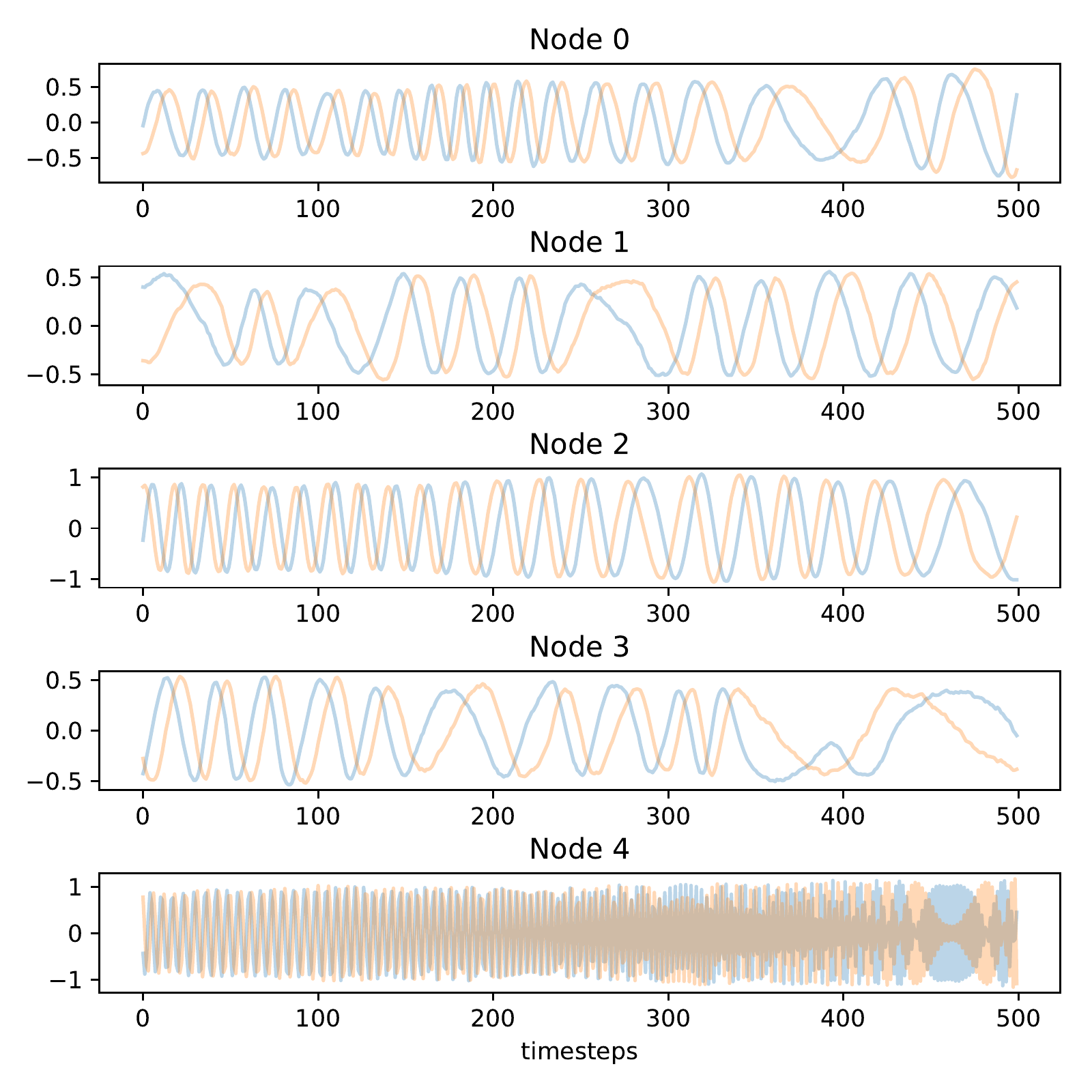}
    \caption{Temporal evolution of the node attributes of a graph in the rotational model of order $10$. Each plot shows the dynamics of the $F=2$ attributes of each node in the graph. 
    }
    \label{fig:dynamics_rot_nodes}
\end{figure}
\begin{figure}[t]
    \centering
    \includegraphics[width=\columnwidth]{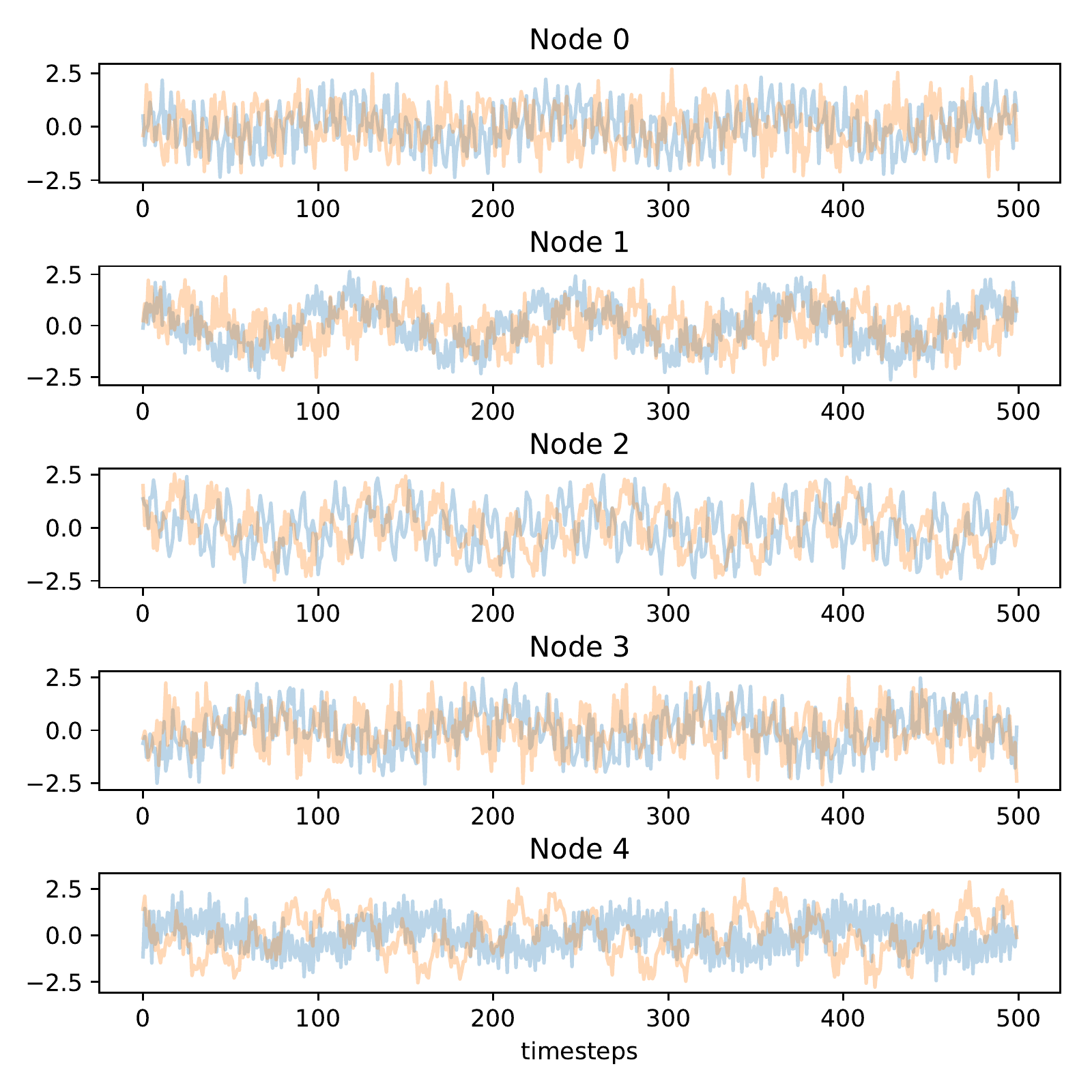}
    \caption{Temporal evolution of the node attributes of a graph in PMLDS($10$). Each plot shows the dynamics of the $F=2$ attributes of each node in the graph. 
    }
    \label{fig:dynamics_pmlds_nodes}
\end{figure}
\begin{figure}
    \centering
    \includegraphics[width=\columnwidth]{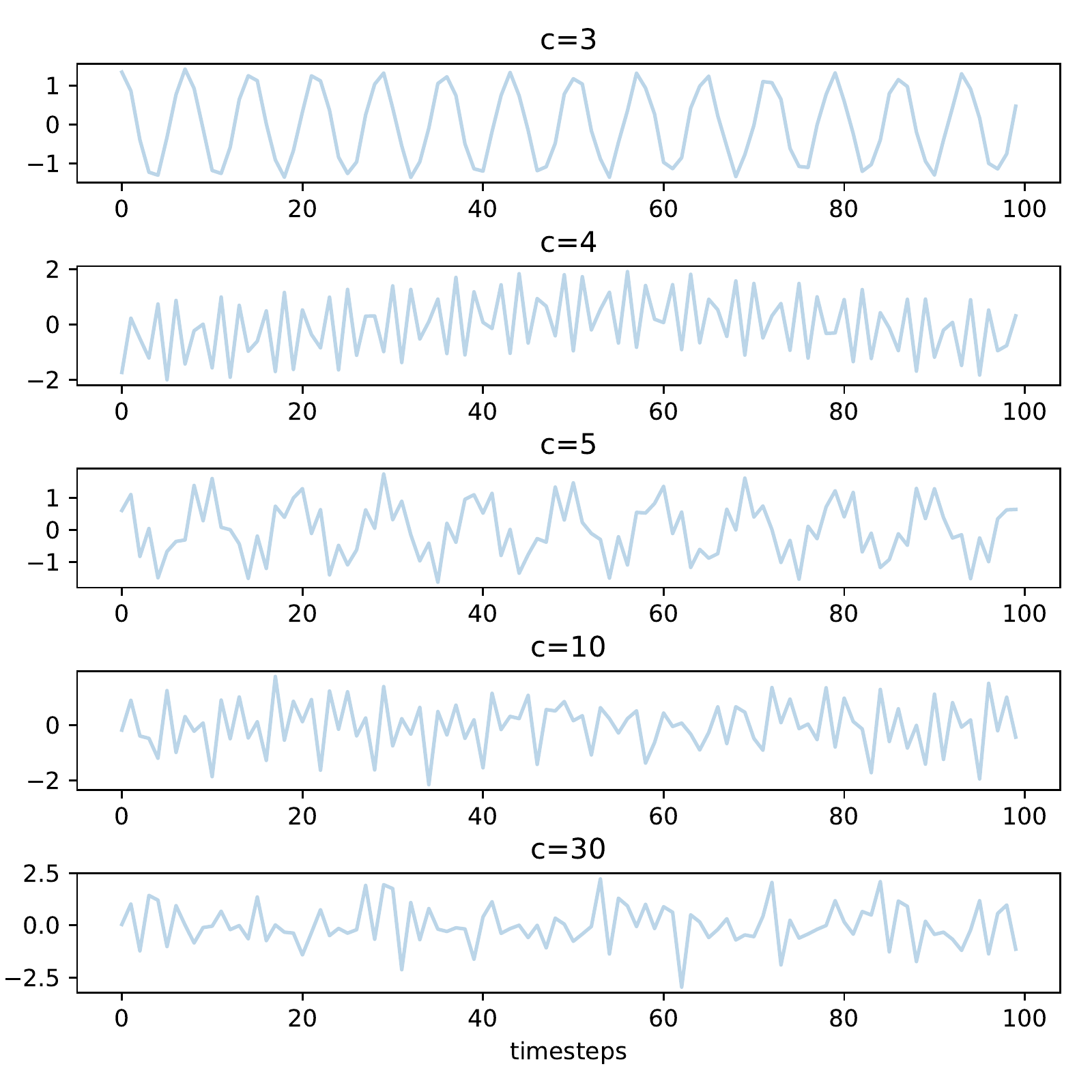}
    \caption{Temporal evolution of the first component of PMLDS, for different values of $c$. As $c$ increases, it becomes more and more difficult to explain the dynamics of the observed coordinate.}
    \label{fig:dynamics_pmlds_full}
\end{figure}
the function $f(\vec x^p_t)$ in \eqref{eq:hidden-AR-model} is taken of the form
\[
   f(\vec x^p_t) = R(\vec x^p_t)\cdot x_t
\]
where $x_{t}\in\R^{N \cdot F}$ and $R(\vec x_t^p)$ is a rotation matrix depending on the regressor.
Matrix $R$ is block-diagonal with blocks $R_n$, $n=1,\dots,N$, of size $F\times F$ defined as:
\[
    R_n(\vec x_t^p) = 
    \left[ 
        \begin{array}{cc}
            \cos(\omega) & \sin(\omega)
        \\ -\sin(\omega) & \cos(\omega)
        \end{array}
    \right]
\]
\[
    \omega = c_n + \alpha\,\cos\left( \sum_{i=0}^{p-1} x_{t-i}(2n-1)+x_{t-i}(2n) \right)
\]
The parameters $\{c_n\}_1^N$ are randomly generated by a uniform distribution in $(-1,1]$, while $\alpha$ is set to $0.01$.

The node attribute matrices are finally obtained by arranging each $x_t$ in a $N\times F$ matrix, and the regression function $f(\cdot)$ can be interpreted as an independent and variable rotation of each node feature (see Figure \ref{fig:dynamics_rot_nodes}).

\subsubsection{Partially Masked Linear Dynamical System}
\label{sec:partial}
we consider a discrete-time linear dynamical system
\begin{equation}
    \label{eq:davies-higham}
    x_{t+1} = R x_{t},
\end{equation}
where $x_t \in \R^c$, and $R \in \R^{c \times c}$ is an orthonormal random correlation matrix computed with the method proposed by Davies and Higham in \cite{davies2000numerically}.

Although the dynamical system \eqref{eq:davies-higham} depends on exactly one previous time step, the partial observation of the first $N\, F< c$ components of $x_t$ results in a dynamical system of order $p \propto (c - N \cdot F)$ \cite{ott2002chaos}.
Similarly to the rotational model, then, node attributes $X_t$ are obtained by reshaping the masked vectors in $\R^{N \, F}$ to ${N\times F}$ matrices (see Figure \ref{fig:dynamics_pmlds_nodes}).
We refer to this setting as a \textit{partially masked linear dynamical system} of complexity $c$ (PMLDS($c$)).

The size $c$ of the original linear system represents an index of complexity of the problem, as the system's memory is dependent on it: given $N$, a higher $c$ will result in more complicated dynamics of $x_t$, and \textit{vice versa} (see Figure~\ref{fig:dynamics_pmlds_full}).  
However, because the evolution of the GGP is controlled by hidden variables that the NGAR model never sees at training time, the problem is closer to real-world scenarios where processes are rarely fully observable. 

\subsection{Graph neural network architecture}
\label{sec:ngar-arch}
\begin{table}[]
    \centering
    \caption{Hyperparameter configuration for the GNN used in the experiments. }
    \label{tab:hyperparameters}
    \begin{tabular}{lc}
        \toprule
         Hyperparameter & Value       \\ \midrule
         Weight for L2  & $0.0005$    \\
         Learning rate  & $0.001$     \\
         Batch size     & $256$       \\
         Early stopping & $20$ epochs \\ 
         \bottomrule
    \end{tabular}
\end{table}
We use the same GNN architecture for both settings.
Input graphs are represented by two matrices:
\begin{itemize}
    \item a binary adjacency matrix, $A \in \{0, 1\}^{N \times N}$;
    \item a matrix representing the attributes of each node, $X \in \R^{N \times F}$.
\end{itemize}

All hyperparameters of $\phi_{nn}$ were found through a grid search on the values commonly found in the literature, using the validation loss as a metric, and are summarised in Table \ref{tab:hyperparameters}. All sub-networks were structured with two layers to provide sufficient nonlinearity in the computation, and in particular $\phi_{conv}$ and $\phi_{dense}$ have been shown in the literature to be effective architectures for processing small graphs \cite{defferrard2016convolutional,simonovsky2018graphvae}.
Our network consists of two graph convolutional layers with 128 channels, ReLU activations, and L2 regularization. Here, we use convolutions based on a first-order polynomial filter, as proposed in \cite{kipf2016semi}, but any other method is suitable (e.g., we could use the edge-conditioned convolutions proposed in \cite{simonovsky2017dynamic} in order to consider edge attributes). The graph convolutions are followed by a gated global pooling layer \cite{li2015gated}, with soft attention and $128$ channels.
The $\phi_{conv}$ block is applied in parallel (i.e., with shared weights) to all $k$ graphs in the input sequence, and the resulting vector sequence is fed to a 2-layer LSTM block with 256 units and hyperbolic tangent activations. Note that we keep a fixed $k=20$ for NGAR, Move, and VAR. 
The output of the LSTM block is fed to a fully connected network of two layers with 256 and 512 units, with ReLU activations, and L2 regularisation.
Finally, the network has two parallel output layers with $N \cdot N$ and $N \cdot F$ units respectively, to produce the adjacency matrix and node attributes of the predicted graph. The output layer for $A$ has sigmoid activations, and the one for $X$ is a linear layer. 
The network is trained until convergence using Adam \cite{kingma2014adam}, monitoring the validation loss on a held-out 10\% of the training data for early stopping. We jointly minimise the mean squared error for the predicted node attributes, and the log-loss for the adjacency matrix. 
For each problem, we evolve the GGPs for $10^5$ steps and test the model on 10\% of the data. 

\subsection{Results}
\label{sec:results}

\begin{table}[]
    \centering
    \caption{Test performance of NGAR on the rotational model of order $p$. We report the full prediction loss (including the regularisation terms), as well as the MSE and log-loss on the node features $X$ and adjacency matrices $A$, respectively. For $A$, we also report the accuracy achieved by the model.}
    \label{tab:results_rot}
    \begin{tabular}{ccccc}
        \toprule
         $p$ & Loss  & MSE ($X$)       & log-loss ($A$)  & Accuracy ($A$) \\ \midrule
         1   & 1.108 & 0.715           & 0.334           & 0.86           \\
         5   & 0.341 & 0.076           & 0.227           & 0.92           \\
         10  & 0.326 & 0.090           & 0.197           & 0.92           \\
         20  & 0.336 & 0.105           & 0.194           & 0.92           \\
         50  & 0.479 & 0.193           & 0.244           & 0.90           \\
         100 & 0.541 & 0.250           & 0.246           & 0.90           \\
         \bottomrule
    \end{tabular}
\end{table}

\begin{table}[]
    \centering
    \caption{Test performance of the NGAR model on PMLDS($c$). We report the full prediction loss (including the regularisation terms), as well as the MSE and log-loss on the node features $X$ and adjacency matrices $A$, respectively. For $A$, we also report the accuracy achieved by the model.}
    \label{tab:results_pmlds}
    \begin{tabular}{ccccc}
        \toprule
         $c$ & Loss  & MSE ($X$)       & log-loss ($A$)  & Accuracy ($A$) \\ \midrule
         11  & 0.200 & 0.018           & 0.154           & 0.95           \\
         15  & 0.278 & 0.041           & 0.195           & 0.92           \\
         20  & 0.283 & 0.038           & 0.204           & 0.92           \\
         30  & 0.341 & 0.081           & 0.222           & 0.90           \\
         60  & 1.366 & 0.983           & 0.345           & 0.85           \\
         110 & 4.432 & 3.950           & 0.418           & 0.82           \\
         \bottomrule
    \end{tabular}
\end{table}

\begin{figure}
    \centering
    \vspace{-.5cm}
    \includegraphics[width=.95\columnwidth]{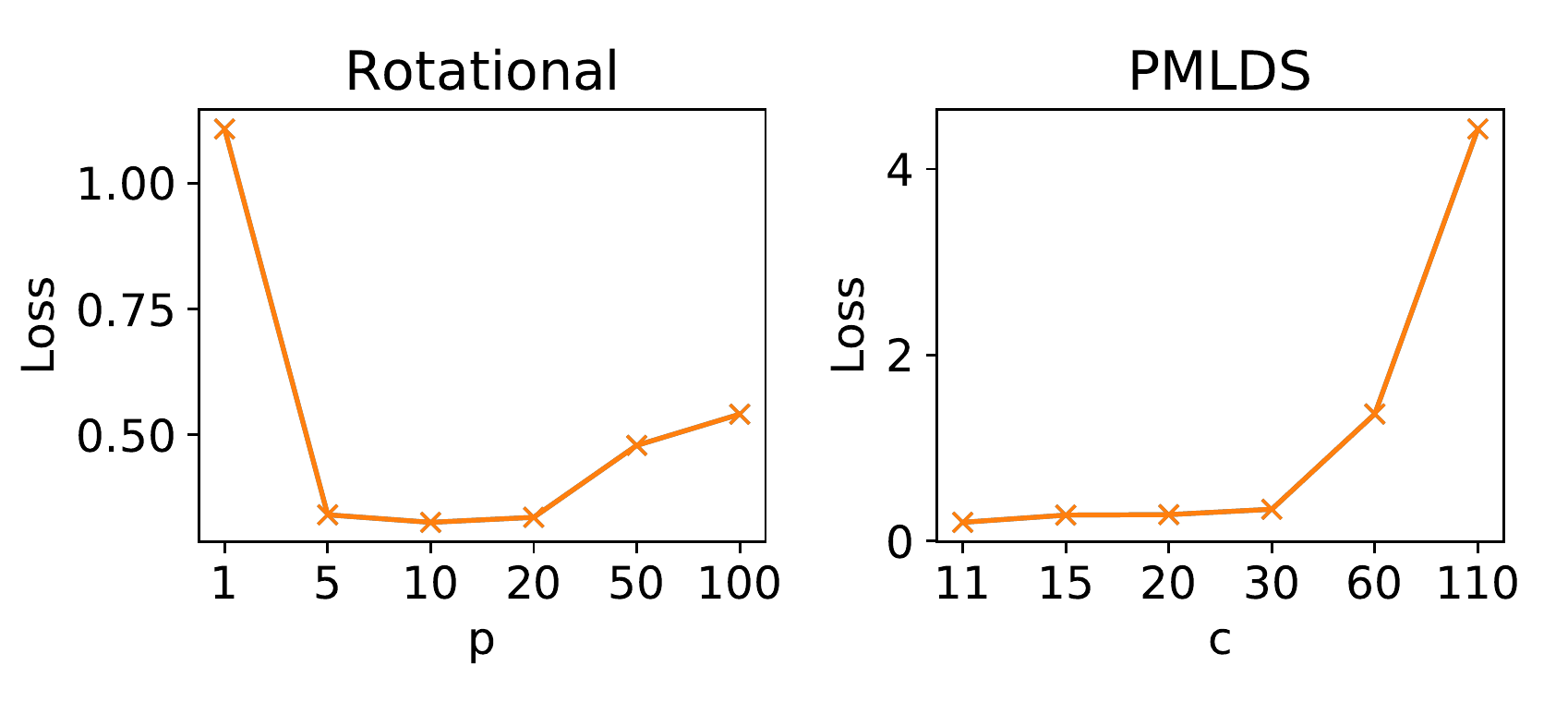}
    \caption{Test loss of NGAR for different levels of complexity on Rotational and PMLDS.}
    \label{fig:loss}
\end{figure}

\begin{figure*}
    \centering
    \includegraphics[width=.95\textwidth]{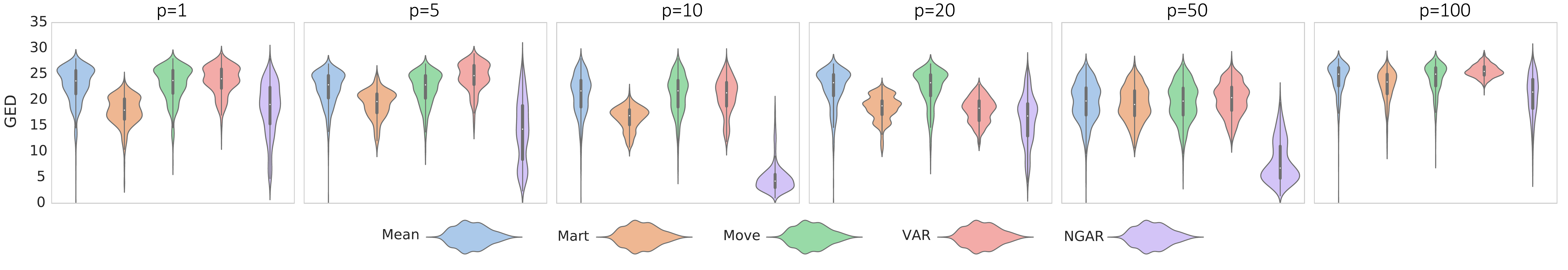}
    \caption{Comparison of \textit{Mean} (blue), \textit{Mart} (orange), \textit{Move} (green), \textit{VAR} (red), and NGAR (purple) on different rotational GGPs. Each plot shows the distribution of the residual GED \eqref{eq:ged-error} between the targets in the test set and the graphs predicted by each model respectively (best viewed in colour).}
    \label{fig:violins_rot}
\end{figure*}

\begin{figure*}
    \centering
    \includegraphics[width=.95\textwidth]{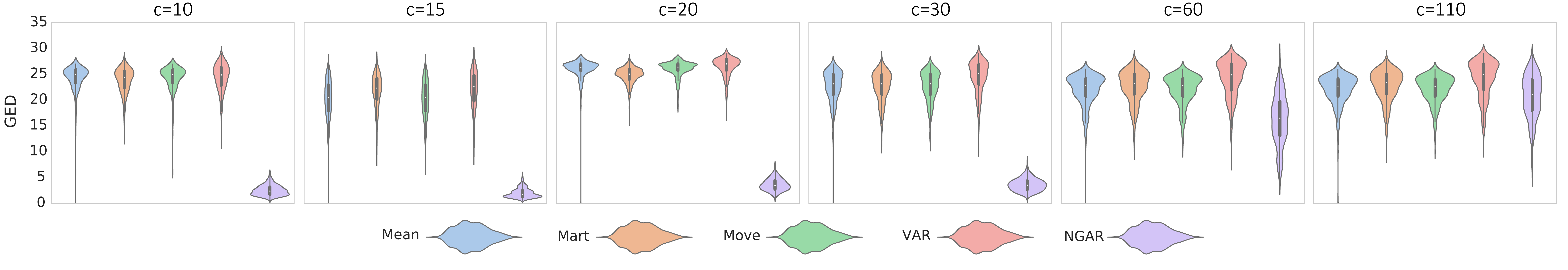}
    \caption{Comparison of \textit{Mean} (blue), \textit{Mart} (orange), \textit{Move} (green), \textit{VAR} (red), and NGAR (purple) on different PMLDS($c$) GGPs. Each plot shows the distribution of the residual GED \eqref{eq:ged-error} between the targets in the test set and the graphs predicted by each model respectively (best viewed in colour).}
    \label{fig:violins_pmlds} 
\end{figure*}

\begin{figure*}
    \centering
    \includegraphics[width=.91\textwidth]{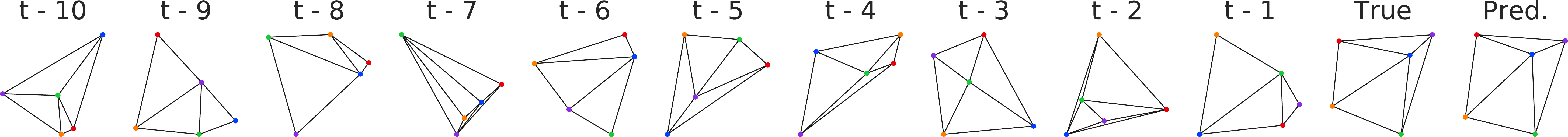}
    \caption{Comparison of a graph predicted by the NGAR model with the ground truth, on PMLDS($30$) (best viewed in colour). Given a sequence of graphs observed up to time $t-1$, it is difficult to intuitively predict the following graph in the sequence (marked ``True'' in the figure). However, the model is able to predict the graph almost exactly (marked ``Pred.'' in the figure).}
    \label{fig:pmlds_seq_target_pred}
\end{figure*}

\begin{figure*}
    \centering
    \includegraphics[width=.95\textwidth]{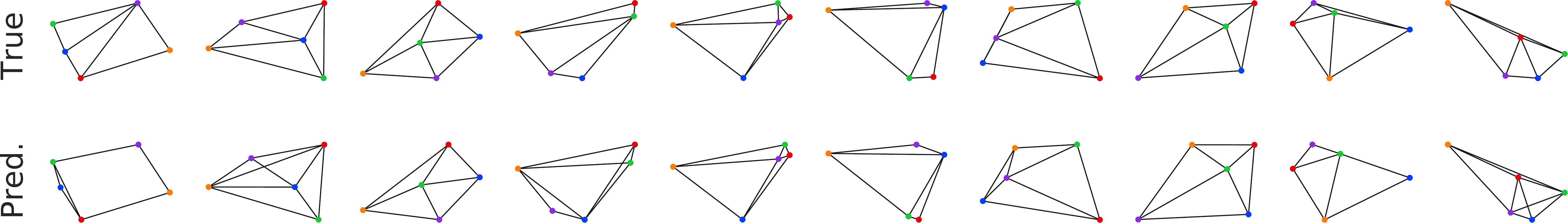}
    \caption{Comparison between randomly selected graphs in the test set of the PMLDS($30$) GGP and the corresponding graphs predicted by the NGAR model  (best viewed in colour). Upper row: samples from the ground truth. Bottom row: graphs predicted by the GNN. }
    \label{fig:predictions_pmlds}
\end{figure*}

Tables \ref{tab:results_rot} and \ref{tab:results_pmlds} report the test performance of NGAR, in terms of prediction loss and accuracy. 
We can see that as $p$ and $c$ increase, the test performance gets worse, which is in agreement with the fact that $p$ and $c$ can be seen as indexes of complexity of the problem. 
In Table \ref{tab:results_rot}, for the rotational GGP with $p=1$ we observe an unexpected high test loss, which might be associated with overfitting. We also see from Figure \ref{fig:loss} that the test performance seems to be consistently better when the complexity of the problem is lower than the order of NGAR, i.e., $p \le k$ and $c - N \cdot F \le k$ (c.f.\ Sec.~\ref{sec:rotational} and Sec.~\ref{sec:partial}).

The second part of the experimental analysis aims at comparing the NGAR method with the baselines in terms of prediction error, assessed as the GED between predicted and ground-truth graphs; Figures \ref{fig:violins_rot} and \ref{fig:violins_pmlds} show a comparison of the baselines on different levels of complexity ($p$ and $c$).
Among the baselines, we cannot clearly identify one with significantly better performance. 
Given the performance of \emph{Mean} and \emph{Mart} for all $p$ and $c$, we see that neither GGP can be modelled as stationary or a martingale.
\emph{Mean} and \emph{Move} performed almost the same in all experiments, while and \emph{Mart} in some cases performed significantly better then the other baselines, e.g., see the rotational GGP with $p=10$.
Despite \emph{VAR} being a method adapted from a widely used model in multivariate AR problems, here it was not able to result in relevant graph predictions. 
On the other hand, NGAR consistently outperformed the baselines on almost all setting. 

Finally, a qualitative assessment of the predictions is presented in Figures \ref{fig:pmlds_seq_target_pred} and \ref{fig:predictions_pmlds}, which highlights a good performance of NGAR even on sequences of graphs that cannot be intuitively predicted. 

\section{Conclusion and future work}
\label{sec:conclusion}

In this paper, we formalised an autoregressive framework for a stochastic process in which the observations are graphs. 
The framework considers a generic family of attributed graphs, in which nodes and edges can be associated with numerical and non-numerical attributes, the topology is allowed to change over time, and non-identified nodes can be considered, as well.
We show that the proposed model for graphs is a non trivial generalisation of the classic multivariate autoregressive models, by introducing the Fr\'echet statistics. 
We proposed also to address the task of predicting the next graph in a graph sequence with a GNN, hence leveraging deep learning methods for processing structured data. 
The GNN implementation proposed here is based on graph convolutional layers and recurrent connections, however when requested by the application, different architectures can be adopted. 

Finally, we performed an experimental campaign in which we demonstrate the applicability of the proposed graph AR model, as well as that the proposed GNN can be trained to address prediction tasks.
The proposed method is compared with four baselines on synthetic GGPs, to rely on a ground-truth for the analysis.
The obtained results are promising, showing that the GNN was able to learn a non-trivial AR function, especially when compared to simpler, statistically motivated baselines. Possible application scenarios to be explored in future works include the prediction of the dynamics in social networks, load forecasting in power distribution grids, and modelling the behaviour of brain networks. 

As a possible extension of this work, we intend to formulate the GNN architecture to work entirely on the space of graphs, without mapping the representation to a vector space. This would require a method for aggregating the graphs of a sequence (e.g., via weighted sum). However, such an operation is currently missing in the literature. 

\bibliographystyle{IEEEtran}
\bibliography{IEEEabrv,main}

\appendix
\subsection{Equivalence between \eqref{eq:ar-multivariate} and \eqref{eq:gar}.}
\label{app:remark:equivalence-proof}

Let $(\mc G,\gdist)$ be the Euclidean space $(\R,\norm{\cdot})$, and $H(a,b)=a+b$, then 
\[
g_{t+1} = H(\phi(\vec g_t^k),\eta) = \phi(\vec g_t^k) + \eta. 
\]
Regarding assumption \eqref{eq:no_bias_graph}, we see that
the integral in \eqref{eq:frechet-mean} (known as Fr\'echet function of $Q$) becomes
\[
\int_{\mc G} \gdist(g,g')^2 dQ(g)
= \int_{\R} \norm{g - g'}^2 dQ(g) 
\]
and 
\begin{multline*}
\int_{\mc G} \gdist(H(\phi(\vec g_t^p),\eta),g')^2 \,dQ(\eta)
= \int_{\R} \norm{\phi(\vec g_t^p) + \eta - g'}^2 \,dQ(\eta)
\\= \int_{\R}\left\{ \norm{\phi(\vec g_t^p) - g'}^2 + \norm{\eta}^2 - 2 \eta(\phi(\vec g_t^p) - g')\right\} \,dQ(\eta)
\\= \norm{\phi(\vec g_t^p) - g'}^2 + \var[\eta] - 2\expect[\eta](\phi(\vec g_t^p) - g').
\end{multline*}
We conclude that \eqref{eq:ar-multivariate} and \eqref{eq:gar} are equivalent, in fact
\[
    \phi(\vec g_t^k) \in \frmu[H(\phi(\vec g_t^k),\eta)]
    \ \Leftrightarrow\  \expect[\eta]=0.
\]
and
\[
    \var[\eta]< \infty \ \Leftrightarrow\  \frvar[\eta]<\infty.
\]

\end{document}